# Development of a General Purpose Sentiment Lexicon for Igbo Language

**Emeka Ogbuju**
Department of Computer Science, Federal University Lokoja, Nigeria
emeka.ogbuju@fulokoja.edu.ng

**Moses Onyesolu**
Department of Computer Science, Nnamdi Azikiwe University, Awka, Nigeria
mo.onyesolu@unizik.edu.ng

## Abstract

There are publicly available general purpose sentiment lexicons in some high resource languages but very few exist in the low resource languages. This makes it difficult to directly perform sentiment analysis tasks in such languages. The objective of this work is to create a general purpose sentiment lexicon for the Igbo language that can determine the sentiment of documents written in the Igbo language without having to translate it to the English language. The material used was an automatically translated Liu's lexicon and manual addition of Igbo native words. The result of this work is a general purpose lexicon – *IgboSentilex*. The performance was tested on the BBC Igbo news channel. It returned an average polarity agreement of 95.75% with other general purpose sentiment lexicons.

## 1. Introduction

Sentiment analysis or opinion mining is a natural language processing task that deals with the determination of positive, negative or neural polarities of texts such as news articles, blogs, reviews or speech presentations at document, sentence or aspect level. Sentiment analysis in English texts had dominated the natural language research because there are many publicly available sentiment lexicons in the language (Liu, 2010; Esuli and Sebastiani, 2006). Though there are publicly available sentiment lexicons in non-English language, the development of a language-specific sentiment lexicon is a resource-intensive task in natural language processing (NLP). Regrettably, the representation of low resource languages is very low in the corpora/lexical development domain. Chen and Skiena (2014) had built sentiment lexicons for 136 languages using graph propagation. However, Igbo language and many other low resource languages across Africa were not included. The objective of this work is to create a general purpose sentiment lexicon for the Igbo language because the Igbo language is among the 2488 endangered languages globally (Palmer and Regneri, 2013); hence there is a need to develop NLP tools for its preservation. As one of the three main languages of Nigeria, it had been included in the working tasks of Windows Operating System in 2009 (Ifeanyi-Reuben *et al.*, 2017) making it open for further computational operations.

There has been active research on the development of general purpose lexica because the use of annotated lexicons is vital in opinion mining. There are existing publicly available general purpose sentiment lexicons in the English language. They include the manually compiled unigrams and the automatically compiled N-grams. The manually compiled unigrams include the MPQA (8000 words annotated with positive, negative, and neutral polarities) by Wilson *et al.* (2005), Liu's opinion lexicon (6800 words categorised into positive and negative) by Hu and Liu (2004), and aFinn lexicon (2500 words rated between -5 to 5 polarities) by Hansen *et al*. (2011). The automatically compiled N-grams include the NRC lexicon (contains 54,129 unigrams, 316,531 bigrams and 480,010 skip bigrams extracted from tweet collection) by Mohammad *et al.* (2013), and Geri lexicons (contains 376,863 unigrams, 922,773 bigrams and 850,074 dependency triples) by Ozdemir and Bergler (2015).

From these lexica, some low resource language specific lexicons had been developed, such as Turkish sentiment lexica – SentiTurkNet, and EmoLex (Hirschberg and Yang, 2017), Bengali and Telugu sentiment lexicons – *SentiWordNet* (Das and Bandyopadhyay, 2010), and Irish sentiment lexicon – *Senti-Foclóir* (Afli *et al.,* 2017). Others include Indonesian (Bojar and Veselovská, 2015), Spanish (Pérez-Rosas *et. al.,* 2012) and Dutch (Smedt and Daelemans, 2012) – all utilizing the

___

WordNet of the given language.

## 2. Materials and Methods

Liu's lexicon was adopted as seed words. Google Translate[1] was used to automatically translate most of the words. The translator was unable to translate a total of 230 positive words and 738 negative words. These were manually interpreted and classified to reflect the native meanings with help from native speakers. Figure 1 shows the opinion lexicons at the translation level with red highlight of the manually translated terms.

```
;;;;;;;;;;;;;;;;;;;;;;;;;;;;;;;;;;;;;;;;;;;;;;;;;;;;;;;;;;;;;;;;;;;
;
; Igbo Opinion Lexicon: Positive
;
; This file contains a list of POSITIVE opinion words (or sentiment words);
; in the Igbo language translated from Liu's Lexicon (2010) by Emeka
; Ogbuju and Moses Onyesolu (2019).
;;;;;;;;;;;;;;;;;;;;;;;;;;;;;;;;;;;;;;;;;;;;;;;;;;;;;;;;;;;;;;;;;;;

a +             adulate         dị ịtụnanya
jupụta          nkwenye         dị ịtụnanya
juru ebe niile  ikpe            oké ochịchọ
ụba             elu             ambitiously
otutu           uru             emeziwanye
ngwa ngwa       bara uru        ike
nwere ike ịnweta n'ụzọ bara uru  mma
na-eti mkpu     uru             amiability
ekwuputara      ihe egwu        amiably
mkpuchi         adventurous     amiable
adalata         akwado          amicability
kwadoro         kwado ya        mma
nabata          akwado          eji obi ụtọ
ọbịbịa          affability      enyi
```

```
;;;;;;;;;;;;;;;;;;;;;;;;;;;;;;;;;;;;;;;;;;;;;;;;;;;;;;;;;;;;;;;;;;;
;
; Igbo Opinion Lexicon: Negative
;
; This file contains a list of NEGATIVE opinion words (or sentiment words)in
; the Igbo language translated from Liu's Lexicon (2010) by Emeka Ogbuju and
; Moses Onyesolu (2019).
;;;;;;;;;;;;;;;;;;;;;;;;;;;;;;;;;;;;;;;;;;;;;;;;;;;;;;;;;;;;;;;;;;;

2-ihu         acridly          ọrịa
2-ihu         acridness        enweghị ihe ọ bụla
ihe ojoo      enwe obi ụtọ     mkpu
wezuga        acrimoniously    egwu
abominable    acrimony         egwu
abominably    ndi mmadụ        dị egwu
kpọrọ asị     n'atụghị egwu    kewapụ
ihe arụ       ogwụ ọjọọ        kewapụrụ
abort         ogwụ ọjọọ        nnweta
aborted       na-eri ahụ       ebubo
abides        ogwụ ọjọọ        ebubo
abrade        na-adụ ọdụ       ekwu
abrasive      ndụmọdụ          enwe nsogbu
ngwa ngwa     na-adụ ọdụ       ọrịa ogwụ
ngwa ngwa     ndụmọdụ          enweghị ahụhụ
ezighi ezi    ndụmọdụ          enweghị aka
enweghị       ikwa iko         altercation
enweghị uche  ikwa iko         ambiguity
```

Figure 1: Sample of Igbo positive and negative opinion lexicons at translation level

The corpora development stages are shown in Figure 2. It progresses from data collection/aggregation to a recursive stage of translation and polarity determination and ends in the pre-processing stage which ultimately normalizes the terms by removal of noise and tokenizes the lexicons by removal of diacritics/accent. The pre-processing tasks and development were carried out using R/RStudio[2] programming.

Figure 2: Igbo sentiment corpora development
(Collection → Aggregation → Lexicons → Translation → Pre-Process → IgboSentilex)

## 3. Results and Discussion

We developed a new sentiment analysis lexicon in the Igbo language known as *IgboSentilex*. It contains 7000 words (2100 positive and 4900 negative) thereby extending Liu's lexicon. The extra 200 words came from a corpus of the Igbo language Bible[3] and sentiment ratings were intuitively determined by the native translators.

A subjective sentiment analysis experiment was done with corpora of the BBC Igbo news channel to test the overall system. The test was carried out with eight (8) corpora from `corpus_ID 01 - 08` containing news categories on entertainment, trending news, movie, etc. A corpus is rated positive at document level if it has more positive words in it and vice-versa for the negative rating. A sample corpus is shown in Figure 3.

---

[1] Google Translate: https://translate.google.com/

[2] R Core Team (2019). R: A language and environment for statistical computing. R Foundation for Statistical Computing, Vienna, Austria. URL https://www.R-project.org/.

[3] Bible Nso (2010). Bible Society of Nigeria. URL: https://www.bible.com/versions/77-igbob-bible-nso

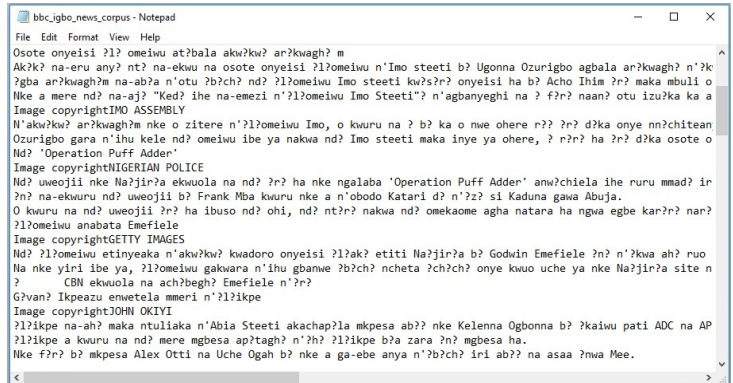

Figure 3: Sample BBC News corpus for analysis

IgboSentilex was compared with one manually compiled unigram (Liu) and one automatically compiled N-gram (NRC). The entire corpus had a 100% agreement except `ID 03` which agreed with Liu and NRC but not IgboSentilex. The overall performance returned an average polarity agreement of 95.75%. Table 1 shows the comparison of the system.

Table 1: Performance comparison

| Sentiment Lexica | Corpus ID/Polarity | 01 | 02 | 03 | 04 | 05 | 06 | 07 | 08 |
|---|---|---|---|---|---|---|---|---|---|
| Liu | Positive | No | Yes | Yes | No | Yes | No | Yes | Yes |
| | Negative | Yes | No | No | Yes | No | Yes | No | No |
| NRC | Positive | No | Yes | Yes | No | Yes | No | Yes | Yes |
| | Negative | Yes | No | No | Yes | No | Yes | No | No |
| IgboSentilex | Positive | No | Yes | No | No | Yes | No | Yes | Yes |
| | Negative | Yes | No | Yes | Yes | No | Yes | No | No |
| Percentage polarity agreement per corpus (%) | | 100 | 100 | 66 | 100 | 100 | 100 | 100 | 100 |
| Average polarity agreement (%): | | 766/8 = **95.75%** | | | | | | | |

## 4. Conclusion

Sentiment lexica are key building blocks for a variety of application, and this contribution for Igbo will help develop technology for the language. Sentiments identified from Igbo native texts may be useful in security, situational relief interventions, document classifications, etc. This work is intended to inspire lexicon translation for other low resource languages in Nigeria and use the results in designing computational NLP tasks. Our future work in line with the corpora development will focus on creating more direct lexicons translated from single root word in the Liu's lexicon and retesting its performance on other corpora apart from news items in the Igbo language.

**Acknowledgement**

We wish to acknowledge the contribution of Emmanuel Chinonso Ezekwem for coordinating the team of native speakers that translated the manual lexicons. We also wish to appreciate the organisers of the Widening Natural Language Processing (WiNLP) workshop for offering the first author a travel grant worth USD1,809 to attend the workshop co-located with the Association for Computational Linguistics (ACL) conference 2019 in Florence, Italy.